\documentclass[10pt]{article} 
\usepackage[preprint]{tmlr}

\usepackage{hyperref}
\usepackage{url}
\usepackage{microtype}
\usepackage{graphicx}

\usepackage{booktabs} 
\usepackage{tabularx}
\usepackage{adjustbox}
\usepackage{multirow}%

\usepackage[ruled,vlined]{algorithm2e}
\usepackage{algpseudocode}%

\usepackage{subcaption}

\usepackage{amsmath}
\usepackage{amssymb}
\usepackage{mathtools}
\usepackage{amsthm}

\title{ \copyright Plug-in Authorization for Human Copyright Protection in Text-to-Image Model}

\author{\name  Chao Zhou \email zc1696190340@mail.ustc.edu.cn \\
      \addr    University of Science and Technology of China
      \AND
      \name Huishuai Zhang \email zhanghuishuai@pku.edu.cn\\
      \addr 	Peking University
      \AND
      \name Jiang Bian \email jiabia@microsoft.com\\
      \addr Microsoft Research 
      \AND
      \name Weiming Zhang \email zhangwm@ustc.edu.cn\\
      \addr University of Science and Technology of China
      \AND
      \name Nenghai Yu \email ynh@ustc.edu.cn\\
      \addr University of Science and Technology of China
}

\def\month{02}  
\def\year{2025} 
\def\openreview{\url{https://openreview.net/forum?id=4o8lIFkpn2}} 

\begin{document}

\maketitle

\begin{abstract}
{This paper addresses the contentious issue of copyright infringement in images generated by text-to-image models, sparking debates among AI developers, content creators, and legal entities. State-of-the-art models create high-quality content without crediting original creators, causing concern in the artistic community and model providers. To mitigate this, we propose the ©Plug-in Authorization framework, introducing three operations: addition, extraction, and combination. Addition involves training a ©plug-in for specific copyright, facilitating proper credit attribution. The extraction allows creators to reclaim copyright from infringing models, and the combination enables users to merge different ©plug-ins. These operations act as permits, incentivizing fair use and providing flexibility in authorization. We present innovative approaches, ``Reverse LoRA'' for extraction and ``EasyMerge'' for seamless combination. Experiments in artist-style replication and cartoon IP recreation demonstrate ©plug-ins' effectiveness, offering a valuable solution for human copyright protection in the age of generative AIs. The code is available at \url{https://github.com/zc1023/-Plug-in-Authorization.git}}
\end{abstract}

 \section{Introduction}\label{sec:intro}

Large foundation models  \cite{gpt3, touvron2023llama, radford2021clip, dalle3, rombach2022high} are trained with extensive, high-quality datasets like The Pile \cite{gao2020pile}, C4 \cite{raffel2020c4}, LAION~\cite{schuhmann2022laion} and other enormous undisclosed data sources, which definitely contain copyrighted human contents. At the same time, these models not only excel at generating content based on user prompts \cite{chatgpt,dalle3, rombach2022high, ramesh2021dalle, ramesh2022dalle2}, but also have the potential of memorizing the exact training data thanks to the huge capacity in their gigantic numbers of parameters \cite{carlini2021extractfromllm, carlini2022quantifying}.

Such training procedure and utilization of AI models have sparked copyright infringement concerns among content providers, artists, and users. A notable instance is the lawsuit filed by The New York Times against OpenAI and Microsoft~\cite{NYT}, alleging the unauthorized use of a vast number of articles for the purpose of training automated chatbots. The lawsuit seeks the destruction of the allegedly infringing chatbots and their associated training data. Similar concerns and legal actions are also emerging in the field of text-to-image generation~\cite{lawsuits}.
 

Indeed, these concerns are well justified as these powerful models could disrupt the existing reward system in creative arts, adding anxiety to the content providers and artist community.  The proficiency of AI in generating artworks that rival human creations is noteworthy, particularly in its ability to replicate characters from major intellectual properties (IPs). For instance, the use of stable diffusion models~\citet{Rombach_2022_CVPR}, combined with controlled generation techniques like ControlNet \cite{zhang2023controlnet}, enables users to effortlessly create well-known characters, such as those from Disney. This ease of replication significantly lowers the barriers to potential copyright infringement, raising concerns about increased piracy risks.

One debating point is whether using copyrighted material to train machine learning models is prohibited by copyright laws. It is known that copyright does not ban all forms of  copying or replication due to the \emph{fair use} doctrine, which allows certain copying and distribution if it can be justified as fair use.   It is not clear whether AI companies can successfully argue that their training procedures fall under this 'fair use' exception in copyright laws~\cite{NYT,lawsuits}. Additionally, academic research is underway to develop methods ensuring AI models do not generate copyrighted concepts, as seen in works like \citet{vyas2023provable}.



In this paper, we step back and advocate to rethink the motivation for enforcing copyright laws. Copyright is a type of intellectual property that intends to protect the original expression of ideas in creative works, which can include literary, artistic, or musical forms~\cite{copyright}.  The foundational goal of copyright laws, as stated in Article I, Section 8, Clause 8 of the U.S. Constitution \cite{USConst},  is ``To promote the Progress of Science and useful Arts, by securing for limited Times to Authors and Inventors the exclusive Right to their respective Writings and Discoveries''.  The primary objective of copyright is to incentivize authors to create new works and to facilitate the dissemination of these works to the public by granting them property rights.  
However, existing generative AI models present challenges in appropriately attributing proper rewards to the copyright holders, which can significantly impact society. Artists, who depend on attribution for recognition and income, may be affected. Additionally, domain experts contributing to knowledge exchange websites like StackOverflow and Quora might hesitate to provide answers if they do not receive reasonable rewards. This situation could backfire on machine learning, as generative models might soon face a shortage of fresh data due to reduced contributions from these sources.

\begin{figure*}[!ht]
\centering
    \includegraphics[width = 0.8\linewidth]{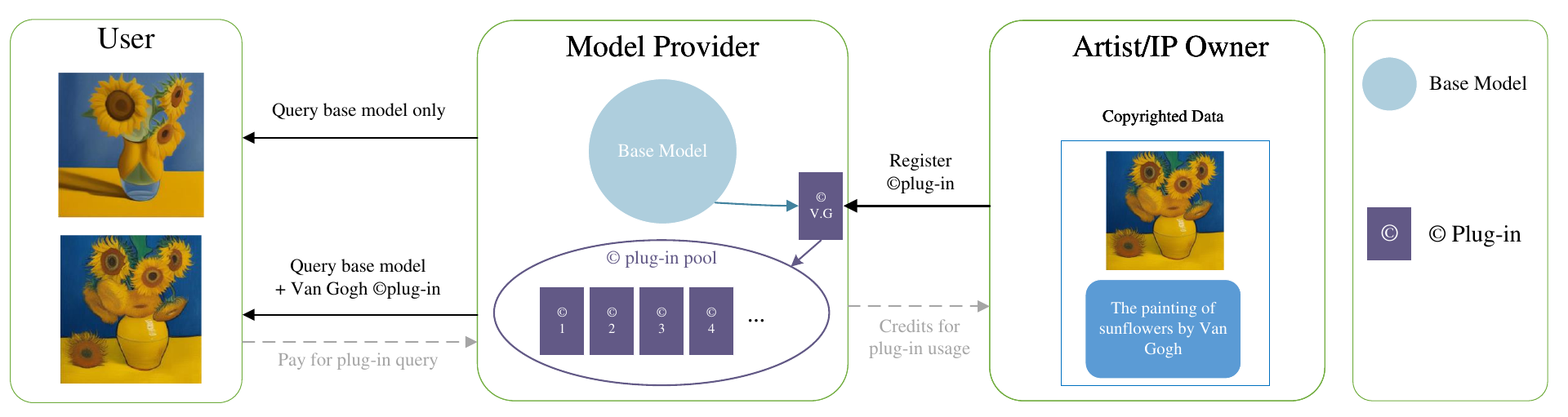}
    \caption{\textbf{\copyright Plug-in Authorization Process.} The authorization process consists of three types of entities: user, model provider, and IP owners (artists). Users can generate copyrighted images only by accessing the relevant plug-in.  The model provider offers services to users, tracks usage of plug-ins, and attributes rewards to the IP owner. The IP owners can achieve authorization by registering their \copyright plug-ins through \emph{addition} or \emph{extraction}. These \copyright plug-ins form a pool where users can get \copyright plug-ins to produce content with the IP owner's authorization.
    }
    \label{fig:market}
\end{figure*}
\begin{figure*}[!ht]
\centering
    \includegraphics[width = 0.8\linewidth]{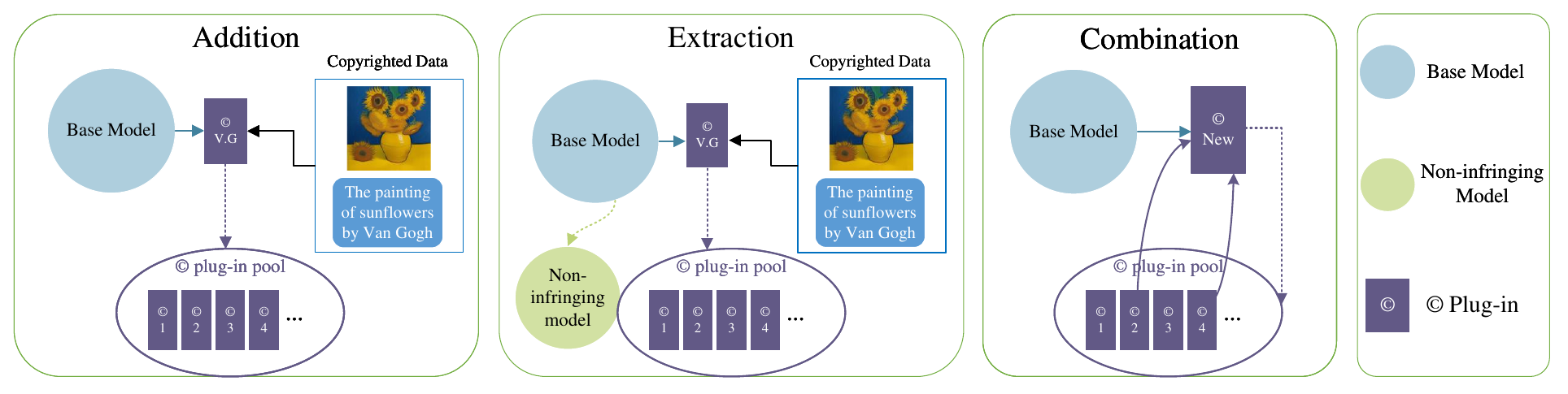}
    \caption{\textbf{Three foundational operations achieving \copyright Plug-in Authorization: \emph{addition}, \emph{extraction}, and \emph{combination}.} The plug-in can be created by addition if the copyrighted work is new to the base model. Meanwhile, the plug-in can be created by extracting from the base model if the copyrighted work is already infringed by the base model.     Once a pool of \copyright plug-ins is constructed, the \emph{combination} operation can merge multiple \copyright plug-ins featuring the generation of multiple concepts and leave a non-infringing model excluding all multiple concepts.
    }
    \label{fig:overview}
\vskip -0.1in
\end{figure*}


To address the attribution challenge in generative AI models, the concept of \emph{Stable Attribution} \cite{stableattribution} has been proposed, aiming to credit artists and share revenue with creators based on their contributions. Specifically, Stable Attribution attempts to trace back an AI-generated image to the most similar examples in the training dataset. However, achieving this with reasonable cost and ensuring fairness is challenging, given the vast size and heterogeneous nature of the training set. Content providers and model owners may have completely different views on the evaluation of the content \cite{NYT}.    While copyright data is unique, its impact on model performance may not be able to be fairly measured, especially when compared to the benefits derived from the vast size of open-domain public data. The value of artistic works might be underestimated by model owners. Addressing these issues is essential to create a balanced ecosystem in generative AI.

In addressing the challenge of practical attribution in generative models, we introduce the ``Copyright Plug-in Authorization'' framework (see Figure \ref{fig:market}), designed to align with existing Intellectual Property (IP) management practices. This framework involves base model providers, like Stability AI, functioning as repositories for copyright plug-ins. Copyright holders, such as artists, can register their works as plug-ins, receiving rewards for their use. End users, in turn, pay for the generation of images involving copyrighted concepts using these plug-ins. This system offers positive incentives for all involved: copyright holders are compensated for their creative contributions, end users can use copyrighted plug-ins without risking infringement, and base model providers profit from plug-in registration and model usage. The framework also facilitates explicit tracking of copyrighted work usage, ensuring a fair and straightforward reward system. By successfully implementing this authorization process, we can enable a more equitable distribution of copyright benefits across the generative model landscape.

{Technically to enable an effective and efficient copyright authorization, the plug-ins, as permits, should be easily created by \emph{addition} if copyrighted works are new to the base models, or by \emph{extraction} if the copyrighted works are already infringed by the base model. Moreover, the plug-ins should be easily \emph{combined}, which allows copyright holders to merge multiple plug-ins into a new one or enables end users to generate images with multiple copyrighted works. Meanwhile, for efficient execution, these operations should be implemented as light adaptations to the base model, e.g., parameter-efficient tuning methods or prompt designs. 

In this paper, we introduce three foundational operations - \emph{addition, extraction}, and \emph{combination} - implemented using the Low-Rank Adaptor (LoRA) method \cite{hu2021lora}. These operations are essential for realizing the Copyright Plug-in Authorization (Figure~\ref{fig:overview} for an overview).

It is noteworthy that Civitai \citep{civitai} represents a commendable attempt to instantiate the \emph{addition} operation, as users can train and share LoRA components to generate corresponding figures.  However, the operations \emph{extraction} and \emph{combination} are currently not publicly available and pose greater challenges. 

The \emph{extraction} operation involves separating the generative model into a non-infringing base model and some copyrighted plug-ins. A conventional approach might involve retraining the model from scratch using only non-infringing data, and then applying LoRA with copyrighted data. However, this method is impractical, if not impossible, due to high training costs and complex data-cleaning processes. 
Alternatively, this paper introduces a  ``Reverse LoRA'' approach to extract a plug-in from an infringing base model. This process begins by capturing the target concept: we LoRA-tune the model on the target concept and then take the negative of the LoRA weights to achieve concept destruction. Then we  fine-tune the LoRA on surrounding contexts to repair the non-infringing model's contextual generation ability.  Finally, we reverse the LoRA  to be the  \copyright plug-in.

The \emph{combination} operation entails merging multiple copyrighted plug-ins into a unified one.  Simply adding these plug-ins together could lead to unpredictable results due to the correlation among copyrighted plug-ins. In this paper, we have successfully developed a method to fuse multiple components. We introduce ``EasyMerge'', a method termed ``data-free layer-wise distillation'' for the combination process. 
Inspired by conditional generation in generative models, we utilize a LoRA component designed to learn the layer-wise outputs of \copyright plug-ins under corresponding conditions. Consequently, the LoRA component can mimic the behavior of these \copyright plug-ins when subjected to the corresponding conditions, effectively achieving the combination of \copyright plug-ins.



Our contributions are summarized as follows:
\begin{itemize}
    \item \textbf{Conceptual contribution: A \copyright Plug-in Authorization framework.} We advocate to solve the problem of copyright infringement in foundation models with a \copyright Plug-in Authorization framework. It can offer a fair and practical solution for the attribution challenge in text-to-image generative models. 
    We further introduce three operations \emph{addition, extraction} and \emph{combination} to instantiate the framework with  efficient human content copyright authorizations.
    
    
    \item \textbf{Technical contribution: A novel ``Reverse LoRA'' algorithm for \emph{extraction}.} It can effectively \emph{extract} copyrighted concepts from the base model,  achieving competitive performance for concept extraction with flexible plug-ins.
    
    \item \textbf{Technical contribution: A novel  ``EasyMerge'' approach for \emph{combination}.}  It is a data-free layer-wise distillation approach, which can effectively and efficiently address the challenge of combining multiple LoRA components.
\end{itemize}

The structure of the paper is as follows. Section \ref{sec:main} introduces the ``\copyright Plug-in Authorization'' framework and delves into the three operations \emph{addition, extraction} and \emph{combination}.  
Section \ref{sec:exp} presents experiments to validate the effectiveness of the proposed operations. Section \ref{sec:discussion} concludes the paper, offering a discussion on the limitations of our work.
 \section{\copyright Plug-in Authorization with  Addition, Extraction and Combination} \label{sec:main}

As detailed in the Introduction, we implement the ``\copyright Plug-in Authorization'' by utilizing the publicly available pretrained diffusion generative model, Stable Diffusion~\citep{Rombach_2022_CVPR}, along with LoRA components~\citep{hu2021lora}. It is important to note that our framework is not confined to specific model structures, thereby facilitating compatibility with other foundation models such as the GPT series~\citep{gpt3}. Additionally, it is capable of synergizing with various light fine-tuning or prompt tuning techniques~\citep{li2021prefix,lester2021power,edalati2022krona,hyeon2021fedpara}. In the subsequent sections, we revisit the fundamentals of diffusion generative models and introduce the three basic operations of the framework, along with our innovative algorithms.



\subsection{Preliminary on Diffusion Generative Model}

Diffusion models~\citep{sohl2015deep,song2020score,ho2020denoising} are probabilistic models designed to learn a data distribution. In the forward pass, Gaussian noises are successively added $T$ times to an image $X_0$, thereby creating a sequence ${X_0, ..., X_T}$ that simulates a Markov process. Conversely, the reverse process trains the model to denoise, effectively emulating the reversal of the Markov Chain. New images are generated by initially sampling random Gaussian noises and then denoising them using the model. Importantly, this process can be conditioned on inputs, such as a prompt text $c$. The denoising process, denoted as ${\Phi}_{(w)}(X_t,c,t)$, is  trained to predict the noise under the textual prompt $c$ at any timestep $t \in [0, T]$, as outlined in \eqref{formula: diffusion objective}.
\begin{equation}
    \mathop{\arg\min}\limits_{w} \ \mathbb{E} _{\epsilon ,X,c,t}\|{\Phi }_{(w)}(X_t,c,t)-\epsilon\|^2
    \label{formula: diffusion objective}
\end{equation}
Recent advancements have introduced \emph{latent diffusion models} as a solution to mitigate the drawbacks associated with evaluating and optimizing models in pixel space, such as low inference speed and high training costs. These latent diffusion models operate within a compressed latent space, exemplified by publicly available models such as the Stable Diffusion Model (SDM), as detailed in \citet{Rombach_2022_CVPR}. The SDM architecture consists of a variational autoencoder (VAE) that maps images to latent space, a U-Net that learns the diffusion process, and a CLIP encoder for text embedding. Our work primarily focuses on the attention structure within the U-Net, which has been identified as the most influential component in diffusion models.

To implement the ``\copyright Plug-in Authorization'', we incorporate three foundational operations into the Stable Diffusion Model (SDM): \emph{addition}, which allows copyright owners to add a plug-in for their works; \emph{extraction}, which enables owners to extract a plug-in from an infringing base model; and \emph{combination}, which permits users to merge plug-ins for multiple copyrighted concepts. The \emph{addition} operation employs LoRA components that are added to SDM's attention matrices, and these are then trained with copyrighted data. While the specifics of the \emph{addition} operation are covered on existing model-sharing platforms like \citet{civitai}, our discussion will primarily focus on the \emph{extraction} and \emph{combination} operations in the subsequent sections.

\subsection{Extraction:  Reverse LoRA }

\begin{figure*}[!ht]
    \begin{center}
    \includegraphics[width = 0.7\textwidth]{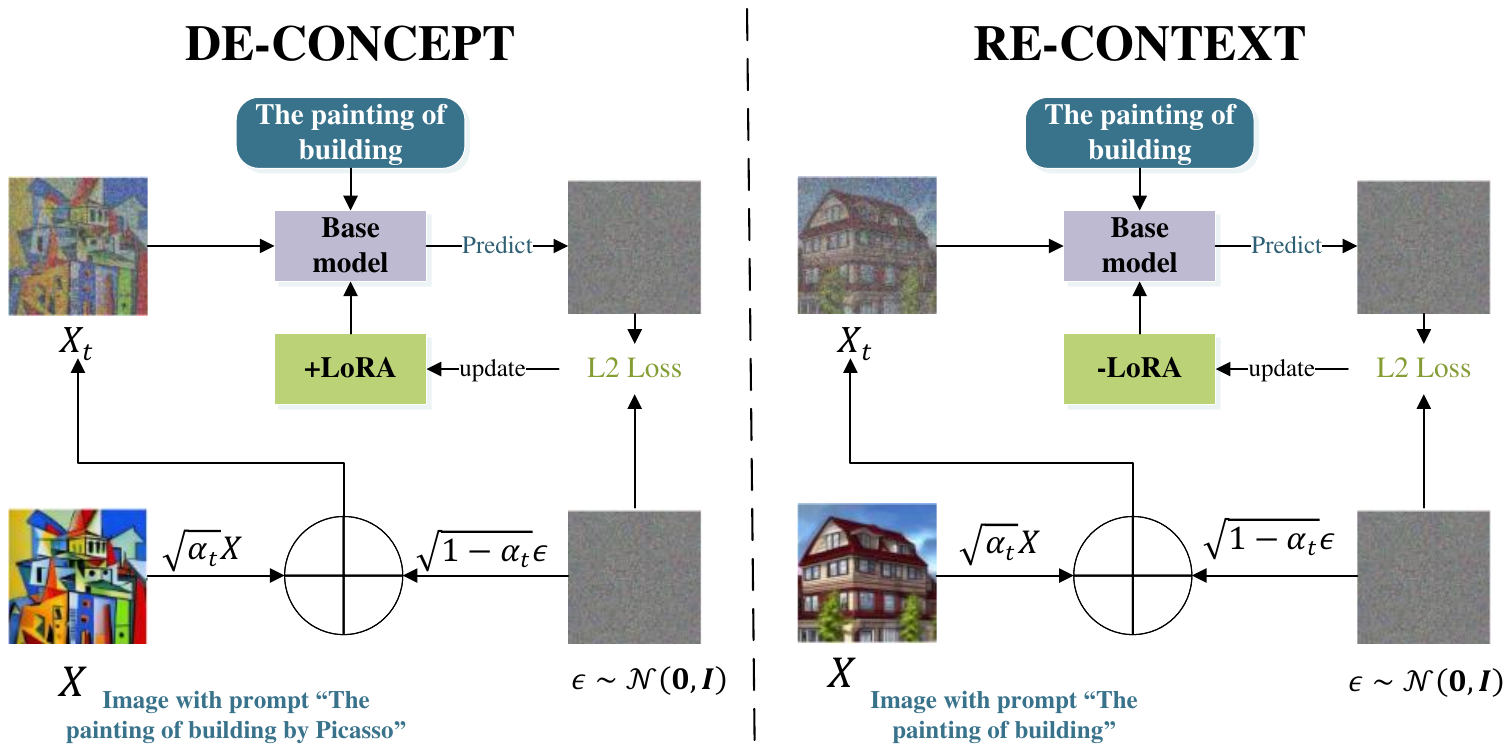}
    
    \caption{\textbf{The method of \emph{extraction}} consists of two steps:\emph{de-concept} and \emph{re-context}. The de-concept step tries to capture the target concept ``Picasso'' by tuning the LoRA component to match copyrighted images with the contextual prompt ``The painting of building''. 
    In the re-context step, we reverse the LoRA (so that successfully forget  ``Picasso'') and then further tune the LoRA with surrounding contextual prompt and non-copyrighted image pairs, to ensure the capabilities of contextual generation. }
    \label{fig:the method of extraction}
    \end{center}
    \vskip -0.2in
\end{figure*}

In our endeavor to achieve \emph{extraction}, we introduce a method known as ``Reverse LoRA''. This approach encompasses two crucial steps to effectively extract the target copyrighted concept while maintaining the ability to generate contextual concepts. The initial step, termed the \emph{de-concept} step, involves removing the target concept from the base model. The subsequent step is a counterbalance to the \emph{de-concept} step, involving relearning the surrounding semantic context, referred to as the \emph{re-context} step.  Figure \ref{fig:the method of extraction}  illustrates these two steps in the context of extracting the target concept ``Picasso''.

\subsubsection{Step1: De-concept}

Our goal is to extract Picasso-related information from the base model to a copyright plug-in using LoRA ($w_L$). This involves identifying the information that represents ``Picasso'' via an alignment process and documenting the changes in model parameters that occur during the alignment process.

Specifically, we align copyrighted image generation, e.g., images of ``the painting of a building by Picasso'', with the  prompt without copyrighted text, e.g., ``the painting of a building'' in the base infringing model. The alignment objective is mathematically expressed as follows:
\begin{equation}
     \mathbb{E}[{\Phi }_{({w})}(\epsilon,c^*,t)] = \mathbb{E}[{\Phi }_{({w+w_L})}(\epsilon,c,t)]
    \label{formula: expected extraction}
\end{equation}
where $\Phi$ is the denoising function, $w$ denotes the original network parameter, $w_L$ is the LoRA component, $c$ is the prompt ``the painting of a  building'',$c^*$ is the prompt ``the painting of a building by Picasso'', $\epsilon$ is the initial noise, and $t$ is the sampling timestep.

To achieve the alignment as defined in \eqref{formula: expected extraction}, we optimize the following objective function with respect to the LoRA parameters  $w_L$, while keeping all other model parameters frozen. The objective function is defined as follows:
\begin{equation}
    \mathop{\arg\min}\limits_{w_L} \ \mathbb{E} _{\epsilon ,X^*,c,t}\|{\Phi }_{(w+w_L)}(X^*_t,c,t)-\epsilon\|^2
    \label{formula: objective}
\end{equation}
where $X^*$ is the copyrighted image (or generated by the infringing model with the prompt ``the painting of building by Picasso''), $X^*_t = \sqrt{\alpha_t}X^* + \sqrt{1-\alpha_t} \epsilon$ is the noisy version of $X^*$,  $c$ is the prompt of ``the painting of building'',   $w$ is the original network and $w_L$ is the LoRA weight. This optimization aims to adjust  $w_L$   so that it effectively captures the desired information related to the target concept.

By incorporating such a LoRA component, the base model can generate Picasso-style images even when the prompts do not explicitly mention ``Picasso''. Hence, the LoRA represents the copyrighted Picasso style, and $w - w_L$ would give us a non-infringing model, which can thought of as an analogy of a `negative LoRA'. However, directly using $w - w_L$ as the non-infringing model compromises its ability to generate images with surrounding context, e.g., ``the painting of a building'', as shown in Figure \ref{fig:result of extraction} in Appendix \ref{app:extraction}. This observation leads us to further tune the LoRA with pairs of images and texts of surrounding semantic context.


\subsubsection{Step2: Re-context}

To mitigate the performance degradation of the non-infringing model when generating images with contextually related prompts, we introduce a re-context step  following the de-concept step. This step involves fine-tuning the LoRA component with images and textual prompts of surrounding contexts, e.g., ``the painting of a building''. To curate the  dataset, we randomly generate images with the base model using the contextual prompt ``the painting of a building'', while leveraging the negative prompt \citep{ho2022classifierfree} ``Picasso'' to steer the generation as far away from the target concept ``Picasso'' as possible.

Specifically, we further optimize ${w}_L$ with the  objective,  
\begin{equation}
    \mathop{\arg\min}\limits_{w_L} \ \mathbb{E} _{\epsilon ,X,c,t}\|{\Phi }_{(w-w_L)}(X_t,c,t)-\epsilon\|^2,
    \label{formula: mem objective}
\end{equation}
where $X,c$ represent the constructed  pairs $(\text{image}, \text{prompt})$ to recover the generation capability of surrounding contexts.

Overall, after the de-concept step, the model $w-{w}_L$ is unable to generate images in the Picasso style, yet it performs well with surrounding prompts thanks to the re-context step.   Therefore, through the \emph{extraction} operation, we obtain a non-infringing model $\tilde{w} = w - {w}_L$ and a \copyright plug-in ${w}_L$.
By incorporating the \copyright plug-in, the model is restored to the original base model $w$, regaining the capability of successfully generating the artworks in the ``Picasso'' style.  The intermediate results of the \emph{extraction} process are visually illustrated in Figure \ref{fig:result of extraction} in Appendix \ref{app:extraction}, showcasing the successful extraction of the targeted copyright, while preserving the model's ability to generate images with surrounding contexts.

\subsection{Combination: EasyMerge}

In this section, we consider the operation \emph{combination}. The combination of existing \copyright plug-ins becomes essential when aiming to generate an image featuring both ``Snoopy'' and ``Mickey'' concepts. 

It is worth noting that simply adding these plug-ins together could yield unpredictable outcomes due to inherent correlations among these plug-ins. To facilitate the combination of multiple copyrighted concepts, we propose a novel approach named  \emph{EasyMerge}. This method employs a data-free, layer-wise distillation technique that only requires plug-ins and corresponding text prompts. Furthermore, with layer-wise distillation, EasyMerge achieves efficient combination in just a few iterations. The versatility of EasyMerge extends beyond the current context, potentially also applicable in other scenarios like continual learning.  

Specifically, we use a new LoRA component $w_L$ to mimic the functionalities of each plug-in that needs to be combined. The objective is defined as follows:  
{\small{
\begin{equation}
        \mathop{\arg\min}\limits_{w_L} \ \sum_{k\in S, j\in S_L} \mathbb{E}_{\epsilon, t}\|{\phi^j_{w-w_L}(\epsilon,c_k,t) - 
        \phi^j_{w-w_{L_k}}(\epsilon,c_k,t)}\|^2,
        \label{formula: objective2}
\end{equation}}}
where $S$ is the set of text prompts to be combined, $S_L$ is the set of layers that are added with LoRA components, and $\phi^j$ is the output of layer $j$'s LoRA component. Similarly to the previous section, $w$ denotes the base model parameter, ${w}_L$ denotes the combined plug-in,  $c_k$ denotes the prompt $k$,  $w_{L_k}$ is the plug-in of context $c_k$, $\epsilon$ is initial noise and $t$ is the sampling timestep of the diffusion process.  The non-infringing model $w-w_L$ is that simultaneously excludes multiple styles related to $c_k$, which is called the \emph{combination} of \emph{extraction}.
Algorithm~\ref{alg:Combination} describes concrete steps of optimizing the objective   \eqref{formula: objective2}.


\begin{algorithm}[h]
\caption{Combination: EasyMerge method}
\label{alg:Combination}
\SetAlgoLined
\KwIn{A set $S$ of indices of plug-ins to be combined, base model $w$, diffusion step $T$}
\KwOut{Combined LoRA $w_L$}

\Repeat{convergence}{
    \For{$w_{L_i}, c_i \in S$}{
        $t \sim \text{Uniform}([1...T])$\;
        $\epsilon \sim \mathcal{N}(0,1)$\;
        AddHook($w_{L_i}$) \tcp*{Capture input $I_{w_{L_i}}^j$ and output $O_{w_{L_i}}^j$ for each layer $j$}
        $I_{w_{L_i}}^j,O_{w_{L_i}}^j \gets \Phi_{w+w_{L_i}}(\epsilon,c_i,t)$ \tcp*{Denoise to obtain features}
        $O_{w_{L}}^j \gets \phi_{{w_{L}}}^j(I_{w_{L_i}}^j)$ \tcp*{Get layer-output through new LoRA}
        $\mathcal{L} \gets \sum_{j\in S_L} \| O_{w_{L}}^j-O_{w_{L_i}}^j\|$\;
        $w_L \gets w_L - \nabla_{w_L} \mathcal{L}$\;
    }
}
\end{algorithm}
 \section{Experiments to Verify  Basic Operations} \label{sec:exp}

As a position paper, we regard our primary contribution as the  proposal of the copyright authorization framework. Nonetheless, we   also aim to validate the practical effectiveness and efficiency of  the proposed basic operations. Given that the \emph{addition} operation has already been well demonstrated by existing practices, we focus on evaluating the \emph{extraction} and \emph{combination} operations. We choose two typical scenarios of copyright infringement: artist-style replication and cartoon intellectual property (IP) recreation.

\subsection{Experiment Setup, Metrics and Baselines}
\textbf{Experiment Setup.} In all  experiments, we fine-tune  the attention component in the U-Net architecture of Stable Diffusion Model v1.5, as described in  \citet{Rombach_2022_CVPR}. 

For the \emph{extraction} operation, we need to generate data with pre-trained models. For the case of extracting a given artistic style, we leverage ChatGPT \citep{chatgpt} to generate 10 common content in paintings. In the de-concept step, for each iteration, we select one of these content to generate 8 images with prompts `` The painting of [content] by [artist]''. Similarly, during re-context step, we select imagery to generate 8 images with prompts `` The painting of [content]'' while using negative prompts ``by [artist]''. 
For the case of extracting a particular IP character, it follows the same procedure as above except that the prompts become ``The cartoon of the [IP character]'' for the de-concept process and  ``The cartoon of the [character]'' for the re-context process, respectively. 
For both the de-concept process and the re-context process, the training consists of  10 iterations, with each iteration 30 epochs. We use a learning rate of 1.5e-4, $T=50$ steps for the diffusion process, and a rank of 40 for LoRA.

For the \emph{combination} operation, we use a learning rate of 1e-3 and a rank value of 140 for LoRA.

\textbf{Metric.} To evaluate the effectiveness of the \emph{extraction} operation, we measure the discrepancy between the set of images generated by the base model and that generated by the non-infringing model after extraction with the same set of prompts. We want to observe a large discrepancy when the prompts are with target concepts while having a small discrepancy with surrounding concepts. This means that the \emph{extraction} operation achieves its goal: the non-infringing model cannot generate images with target concepts but can generate high-quality images with  surrounding prompts. 

We acknowledge that for image generation tasks, the ultimate evaluation criterion is human judgment. Therefore, we provide the generated images from various scenarios for readers' assessment. Nevertheless, to reduce costs and facilitate comparisons with existing approaches, we also employ an objective metric known as the \emph{Kernel Inception Distance} (KID)~\citep{binkowski2018demystifying} to quantify the aforementioned discrepancy. KID is akin to the \emph{Fr\'echet Inception Distance} (FID)~\citep{heusel2017gans} but is considered to exhibit less bias and possess asymptotical normality. Moreover, we also employ the Learned Perceptual Image Patch Similarity (LPIPS)~\citep{zhang2018perceptual} to quantify the discrepancy of artistic style artworks. LPIPS is a robust measurement tool that effectively captures differences in human perception between two images, offering a  comprehensive evaluation of stylistic variations in generated artworks.

\textbf{Baseline}. We compare our \emph{extraction} operation with the concept ablation approach~\citep{kumari2023ablating} and  Erased Stable Diffusion (ESD) \citep{gandikota2023erasing}, which achieve  concept removal by aligning  latent representations of target concepts with those of anchor concepts. 

In general, we find it hard to compare the results with existing methods because of the complex setups in image generation, e.g., the tuning steps and the trade-off between removing the target concept and keeping the surrounding concept. Therefore we take a conservative approach and only consider the generation with similar scenarios and the same metric as in the original paper. 



\subsection{Extraction and Combination of Artists' Styles}

\begin{figure}[!htb]
\centering
    \begin{subfigure}{0.45\linewidth}
        \includegraphics[width = \linewidth]{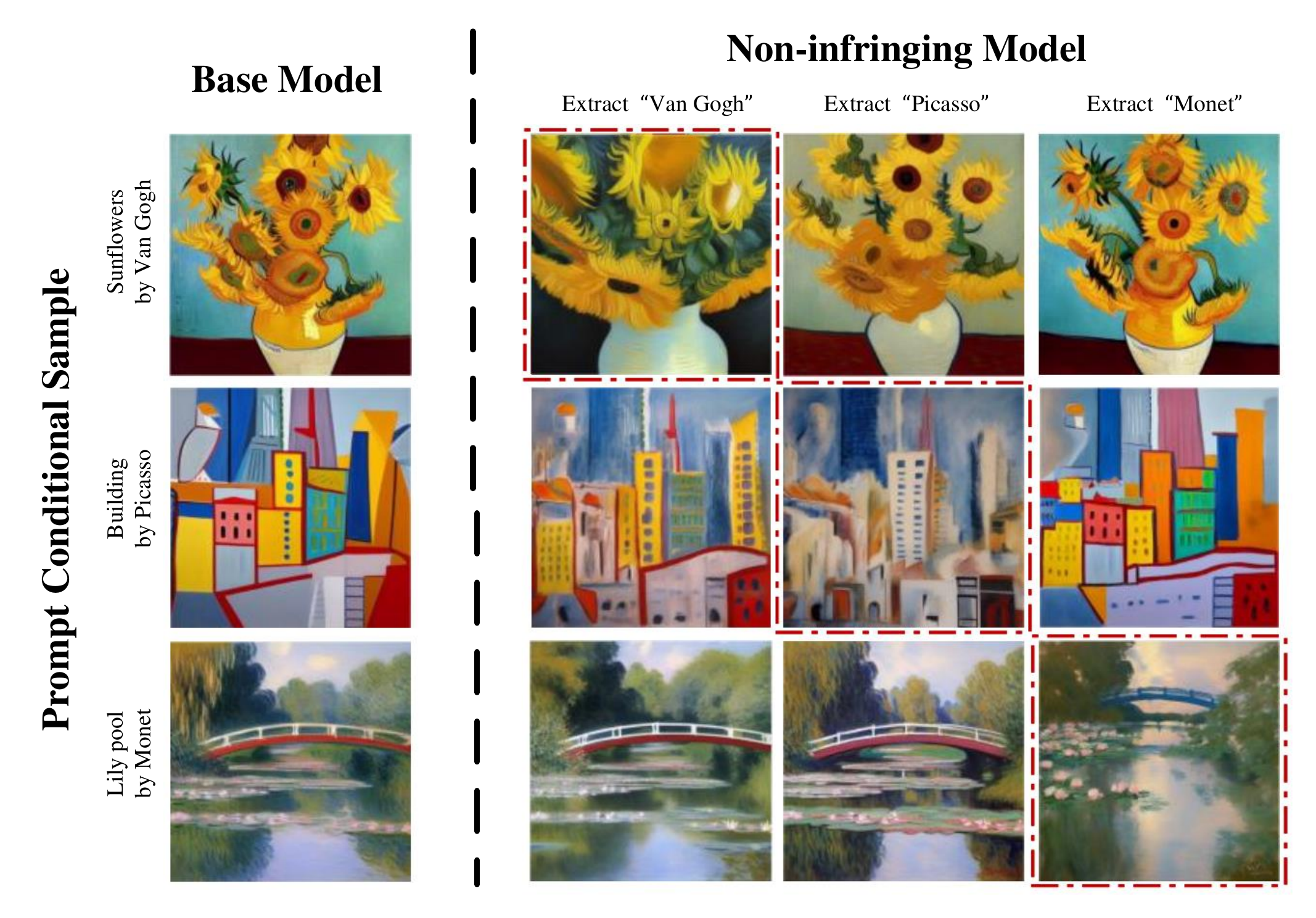}
        \caption{ Results of \emph{extraction} in style replication}
        \label{fig:Results of extraction in style transfer}
    \end{subfigure}
    \hfill
    \begin{subfigure}{0.45\linewidth}
        \includegraphics[width = \linewidth]{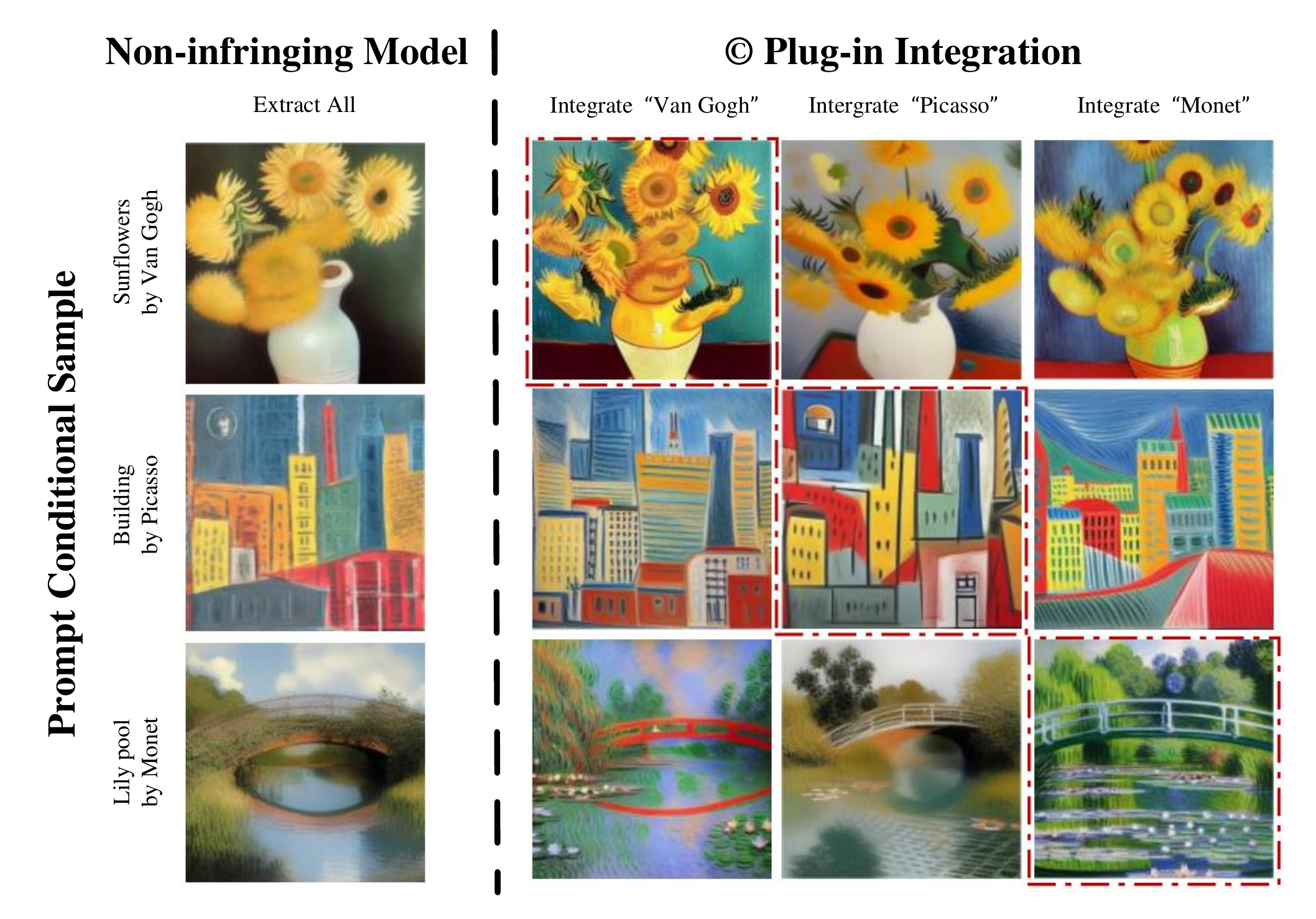}
        \caption{ Results of \emph{combination} in style replication}
        \label{fig:Results of combination in style transfer}
    \end{subfigure}
    \caption{\textbf{Results of style replication}. In Figure (a), we show samples from different non-infringing models in each column. Each non-infringing model exhibits a deficiency in one style generation ability, with all other style generation capabilities remaining unaffected.
    In Figure (b), we present samples generated after integrating certain \copyright Plug-ins in each column. Each of these \copyright Plug-ins serves to exclusively restore the generation of one particular style, while the generation of other styles continues to exhibit diminished performance.}
    \label{fig:results of style transfer}
\end{figure}

\textbf{Extraction.} We extract artist styles from the Stable Diffusion V1.5, referred to as the ``base model''. We consider three renowned artists: (1) Vincent van Gogh, (2) Pablo Ruiz Picasso and (3) Oscar-Claude  Monet. The results of individual style extractions are visually presented in Figure \ref{fig:results of style transfer}(a). These images showcase outputs generated by both the base model and the non-infringing model, encompassing both the target styles and surrounding styles.

In Figure \ref{fig:results of style transfer}, images within red boxes represent the target styles, while the rest images  embody surrounding styles.  A notable contrast is observable between the images within the red boxes and those generated by the base model. However, the images representing surrounding styles exhibit a substantial similarity to those generated by the base model.  This demonstrates the success of the \emph{extraction} operation in isolating the target style from the base model while preserving the quality of images with surrounding styles.

\begin{table}[!ht]
\caption{\textbf{Quantitative comparison with baselines in artist-style extraction.} Compared to Concepts-Ablation, ours extracts the target style more thoroughly, and compared to ESD, ours enjoys less damage to surrounding styles.}
\label{KID $X 10^3$ of non-infringing model}
\begin{center}

\begin{small}
\begin{sc}
\scalebox{0.8}
{
\begin{tabular}{cccc}
\multicolumn{1}{c}{\bf Metrics} &\multicolumn{1}{c}{\bf Methods} &\multicolumn{1}{c}{\bf Target style $\uparrow$}  &\multicolumn{1}{c}{\bf Surrounding style $\downarrow$}\\
\midrule
\multirow{2}*{KID$\times 10^3$} &Extraction (Ours)        &\textbf{187} &32 \\
~ &Concepts-Ablation         &42 & 12\\
\\
\multirow{2}*{LPIPS} &Extraction (Ours)      &0.31 &  \textbf{0.14}\\
~ &ESD       &0.38 & 0.21\\

\end{tabular}
}
\end{sc}
\end{small}
\end{center}
\vskip -0.1in
\end{table}

In Table \ref{KID $X 10^3$ of non-infringing model}, we employ quantitative metrics to evaluate the effectiveness of the \emph{extraction} operation in comparison to baseline methods. Our method demonstrates notable improvements, as indicated by the KID metric increasing from 42 to 187 for the target style when compared to Concepts-Ablation \citep{kumari2023ablating}. This increase indicates an enhanced removal of the target style. Additionally, in a comparative assessment with the Erasing method \citep{gandikota2023erasing}, our method achieves a reduction in LPIPS from 0.21 to 0.14 for surrounding styles. This reduction implies less degradation of the surrounding artistic styles, affirming our method's ability to preserve the quality of generated images when using surrounding style prompts.

\textbf{Combination.} 
In this part, we show the effectiveness of  the \emph{combination} operation for extracting multiple artist styles and then  adding them back with corresponding plug-ins. 

Given  three artistic styles of Van Gogh, Picasso, and Monet and their \copyright plug-ins, we first extract these three styles from the base model, producing a non-infringing model, which is illustrated in  the leftmost column of Figure \ref{fig:results of style transfer}(b).  
Notably, all the images generated by the non-infringing model significantly differ from those generated by the base model in the leftmost column of  Figure \ref{fig:results of style transfer}(a). This underscores the efficacy of the combination of multiple \emph{extraction} operations.


We then individually integrate each style copyright plug-in into the non-infringing model. The images highlighted within red boxes represent the target style achieved after integrating the respective copyright plug-in. 
Notably, the target style images after integration are distinctly different from those produced by the non-infringing model, showing a closer resemblance to the images generated by the base model. This observation indicates that the  copyright plug-in  can reinstate  the model's ability to create artworks in the target styles, without infringing upon the copyright restrictions associated with other artistic styles. 

\subsection{Extraction and Combination of Cartoon IPs}

In the context of intellectual property (IP) recreation, we demonstrate the capabilities of our framework through both \emph{extraction} and \emph{combination} operations. Specifically, Figure \ref{fig:results of IP recreation} displays the outcomes of extracting three iconic IP characters: Mickey, R2D2, and Snoopy. The images framed in red boxes were generated by the non-infringing model using prompts specific to the target IP, after extraction. These images significantly deviate from those produced by the base model. In contrast, images outside the red boxes, which represent other IPs, show a resemblance to those generated by the base model, indicating the targeted nature of the extraction process.

\begin{figure}[!ht]
    \includegraphics[width = \linewidth]{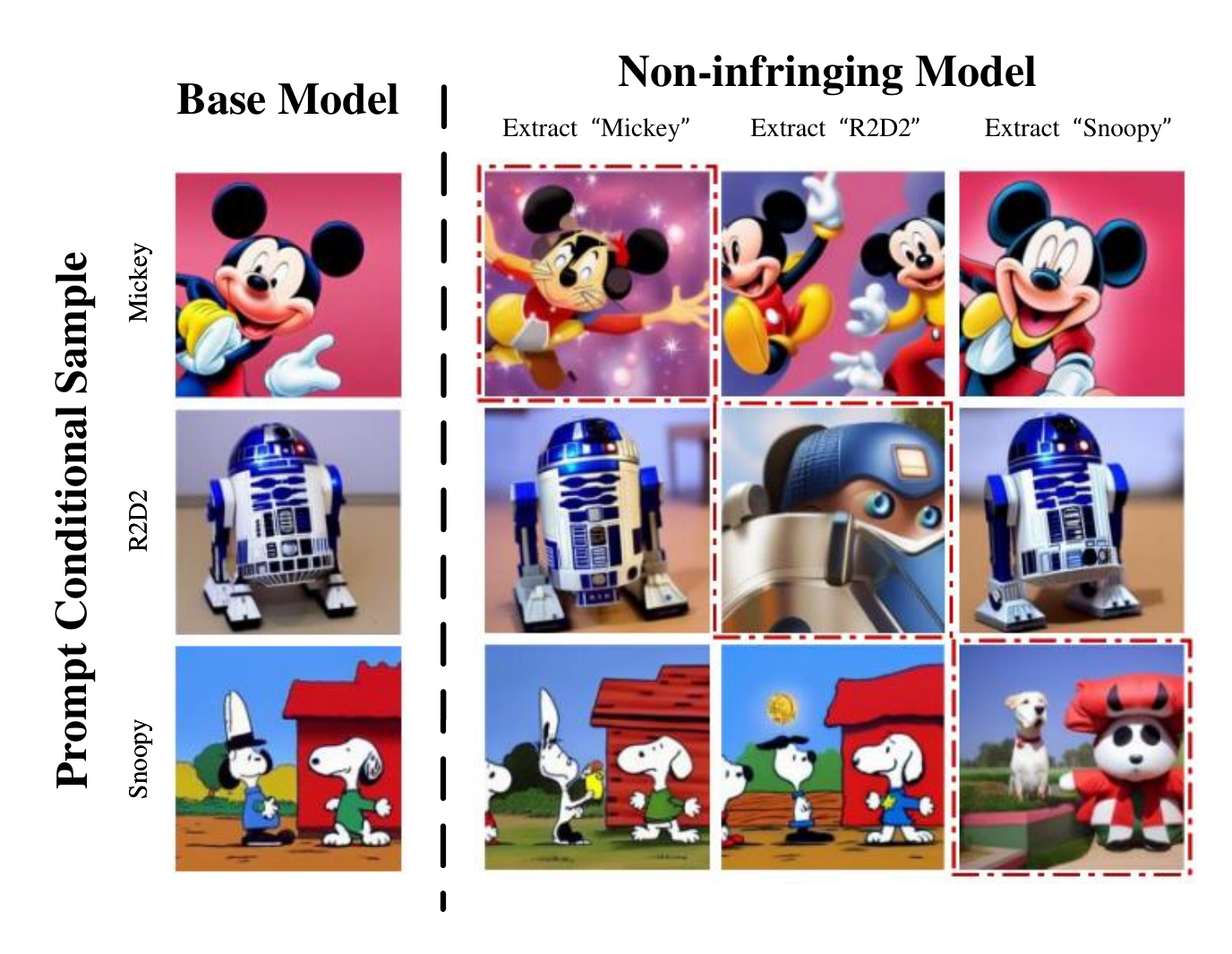}
    \caption{\textbf{Results of IP Recreation.} Each column on the right represents the output of a distinct non-infringing model. We successfully extract the unique IPs of Mickey, R2D2, and Vader independently, preserving the generation of other IPs.
    }
    \label{fig:results of IP recreation}
\end{figure}

Our approach to IP extraction effectively isolates the specified IP, ensuring that the generation capabilities for other IPs remain intact. The efficacy of our extraction method in the realm of IP recreation is quantified in Table \ref{Table:Quantitative comparison in IP recreation}, where we document a notable improvement, i.e., approximately a 2.6-fold increase in the Kernel Inception Distance (KID) metric for the targeted IP, while the KID metrics for other IPs remain relatively stable.

Additionally, our comprehensive large-scale experiments demonstrate that the extraction process does not impact the model's ability to generate non-IP-related content. The performance of the model after extraction remains consistent with routine or everyday use cases, as further detailed in the appendix \ref{app:MScoco}. This ensures that while the model respects copyright constraints by effectively removing or isolating specific IPs, it does not compromise on its general utility or the breadth of its creative outputs.

\begin{table}[h]
\caption{ \textbf{Quantitative comparison in IP recreation.} We increase the KID of the target IP about 2.6 times compared with Concepts-Ablation while keeping the KID of the surrounding IP on par.}
\label{Table:Quantitative comparison in IP recreation}
\begin{center}
\begin{small}
\begin{sc}
\scalebox{0.8}{
\begin{tabular}{cccc}
\multicolumn{1}{c}{\bf Metrics} &\multicolumn{1}{c}{\bf Methods} &\multicolumn{1}{c}{\bf Target IP $\uparrow$}  &\multicolumn{1}{c}{\bf Surrounding IP $\downarrow$}\\
\midrule
\multirow{2}*{KID$\times 10^3$} &Extraction (Ours)        &\textbf{131} & 17 \\
~& Concepts-Ablation          &50 & 15\\
\end{tabular}
}
\end{sc}
\end{small}
\end{center}
\end{table}


Furthermore, we demonstrate the  combination of multiple IP  \copyright plug-ins, as illustrated in Figure~\ref{fig:results of IP recreation combination}. The initial image is produced by the non-infringing model after extracting Mickey Mouse and Darth Vader, where the IPs are hard to recognize. Subsequent images, the second and third, are created after individually adding the Mickey and Vader \copyright plug-ins, which distinctly feature the respective IP. The final image is generated upon adding the combined \copyright plug-in, successfully displaying both IPs. This procedure underscores the plug-in's efficacy in selectively and collectively restoring the model's ability to generate IP-specific content.

\begin{figure}[!ht]
    \includegraphics[width = \linewidth]{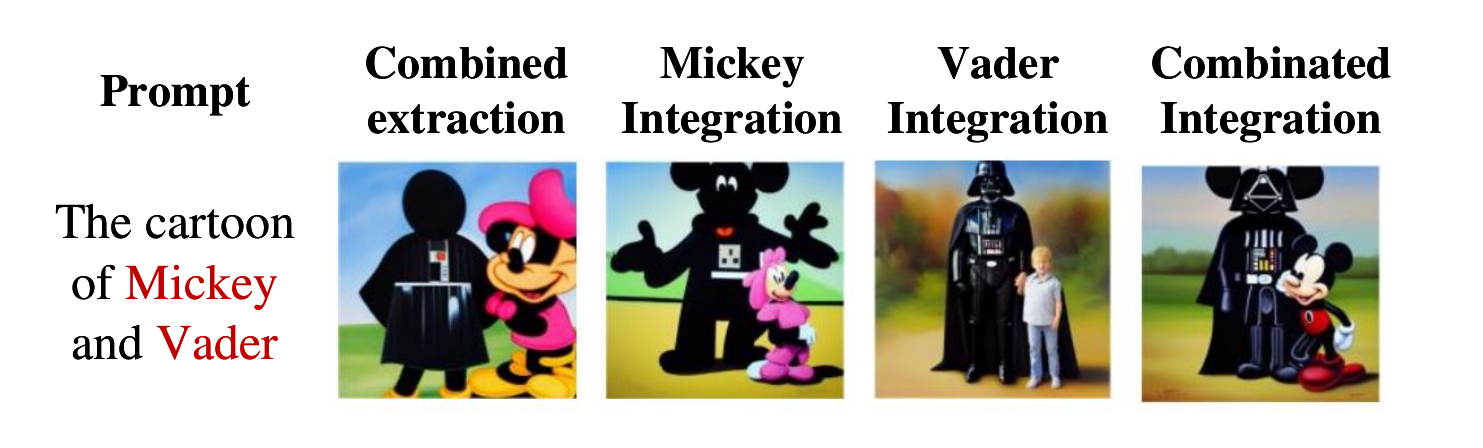}
    \caption{\textbf{IP Recreation in a single image.} 
We can integrate \copyright plug-in into the non-infringing model to generate either Mickey or Vader in a single image or integrate the combined \copyright plug-in to generate both of them.
    }
    \label{fig:results of IP recreation combination}
\end{figure}


These results indicate the efficacy of combining multiple extractions, where the non-infringing model's ability to produce images themed with either Mickey Mouse or Darth Vader is disabled. With the integration of the respective copyright plug-in, the model regains the capability to create content related to the specific IP. 
 \section{Related Work} \label{sec:related}

To position our work in the vast literature, we review related work through two perspectives: scope and technique. It is worth noting that some of the literature touches both sides and we organize them in a  way most related to ours.

\subsection{Scope Related: Copyright, Data Contribution and Credit Attribution}

Recent text-to-image generative models are trained with large-scale datasets \citep{schuhmann2022laion, liu2022taisu}, which cannot be guaranteed free of copyrighted data. At the same time, the state-of-the-art models are capable of generating high-quality and valuable creative images comparable to human creators or even memorizing the data points in the training set \citep{carlini2023extracting}, which arouses copyright concerns about the training data and brings anxiety to the artist community. 

Numerous efforts have been made for copyright protection of training data  \citep{zhong2023copyright}. A direct approach is removing the copyrighted images from the training set, which may involve cumbersome costs due to the size of the training sets and may significantly degrade the model performance \citep{feldman2020does}.  Another direct approach is post filtering, refusing to generate images with copyrighted concepts, e.g., \cite{schramowski2022safe} proposes \emph{Safe Latent Diffusion } to guide latent representation away from target concepts in the inference process, which nonetheless can be bypassed by a user with access to the model \citep{rando2022red}. As an example, OpenAI Dall$\cdot$E3 \citep{dalle3} declines requests for generating an image in the style of a living artist and promises that creators can also opt their images out from training of future image generation models.  Many papers discuss the idea of concept removal, which will be reviewed in later section.

\citet{shan2023glaze} propose \emph{Image Cloaking} that suggests adding adversarial perturbations before posting artistic works on the internet so as to make them unlearnable for machine learning model, which has been pointed out to be hard to defend against future learning algorithms \citep{radiya2021data}. 

Theoretically, \cite{bousquet2020synthetic, elkin2023can} connect the copyright protection of training data with the concept of differential privacy and discuss their subtle differences. \cite{vyas2023provable} further formulate the copyright problem  with a \emph{near free access} (NAF) notion to bound the distance of the generative distributions of the models trained with and without the copyrighted data. 

Our paper distinguishes largely from all previous works as we do not try to prohibit generating copyrighted concepts but instead we introduce a copyright authorization for the generative model to reward the copyright owners with fairness and transparency. From this aspect, our paper is also related with literature of monetizing the training data \citep{vincent2021deeper, vincent2021data, li2022measuring, li2022all} or attributing credits for the generative contents \citep{stableattribution}, but we establish a very distinct way to reward the authorship.  

Heated discussion is also around the copyright for AI generated artwork \citet{franceschelli2022copyright, abbott2022disrupting}. The Review Board of the United States Copyright Office has recently refused the copyright registration of a two-dimensional AI generated artwork entitled “A Recent Entrance to Paradise''. However,  \cite{abbott2022disrupting} argues for giving the copyright to AI generated works, which will encourage people to develop and use creative AI, promote transparency and eventually benefit the public interest. 

\subsection{Technique Related: Concept Removal, and Negative Sampling}

Our \emph{extraction} operation is closely related with the \emph{concept removal} for generative models. \cite{gandikota2023erasing, kumari2023ablating} remove target  concepts by matching the generation distribution of contexts with target concepts and that of contexts without target concepts. \cite{zhang2023forget} forget target concepts by minimizing the cross attention of target concepts with that of target images. \citet{heng2023selective} leverage the reverse process of continual learning to promote the controllable forgetting of target contents in deep generative models.


We note that negative sampling~\citep{ho2022classifierfree} can also prevent generating  certain concepts. Specifically, end users can set conditional context and negative context to guide the diffusion process to generate images conforming the conditional context while being far away from the negative context. Only negative sampling cannot stop copyright infringing generation because the  contexts are  set freely and adversarially by end users.  

In contrast, for a specific copyrighted concept, our \emph{extraction} operation takes an ``Reverse LoRA'' approach to disentangle the base model into two part: non-infringing base model and the plug-in LoRA component for copyrighted concept. Specifically, we use negative sampling to generate non-infringing images, which serves as training data for copyright plug-in. From the aspect of parameter efficient fine-tuning, our paper is related with literature \citep{alaluf2022hyperstyle,ruiz2023dreambooth,gal2022image,hu2021lora,huang2023lorahub}.

Our \emph{combination} operation is related with the widely studied ``knowledge distillation''  \citep{liang2023less,lopes2017data,sun2019patient,hinton2015distilling,fang2019data}, but entails large difference from previous work. We combine multiple copyright plug-ins that are LoRA components for different targets, and we take data free approach due to practical constraint.

 \section{Discussion, Open Questions and Limitations} \label{sec:discussion}

The growing concerns regarding generative AI models stem from their capacity to produce copyright-infringing content. This issue becomes more pronounced as state-of-the-art models continue to improve the quality of generated images, often without adequately acknowledging the contributions of human content creators. In response, we propose the ``Copyright Plug-in Authorization'' framework to address these societal worries, drawing inspiration from the purpose of copyright law. Our approach demonstrates that copyrighted data can be incorporated into LoRA plug-ins, enabling straightforward tracking of usage and equitable distribution of rewards.

A key challenge for this framework is the efficient management of a large number of plug-ins, which is essential to ensure user-friendly access to specific generations. Moreover, updating the base model poses another challenge, as retraining the entire suite of plug-ins can be costly, raising the issue of ensuring backward compatibility. One limitation of our current research is the potential degradation in the performance of the non-infringing model due to a large number of extraction operations, a factor that has yet to be thoroughly investigated.

\bibliography{tmlr}
\bibliographystyle{tmlr}

\newpage
\appendix

\section{Appendix}
\subsection{Intermediate results of extraction}\label{app:extraction}

We present an intricate analysis of the \emph{extraction} process, elucidating distinct phases that delineate the evolutionary trajectory of the non-infringing model's performance. Following the de-concept step, the model experiences a transient phase marked by a temporary impairment in its ability to generate semantically meaningful images. Subsequently, the re-context step engenders a noticeable restoration, enhancing the model's proficiency in generating semantically rich images. Importantly, despite this recuperation, the model retains its inherent limitation – the incapacity to generate artwork in the distinctive style synonymous with Picasso. This observation underscores the success of the de-concept step, wherein the LoRA component effectively captures and removes the target concept, leading to the temporary impairment in the non-infringing model. The subsequent re-context step rectifies this performance decrease without reintroducing any information about the target concept. In essence, the \emph{extraction} process successfully achieves the targeted concept extraction.

\begin{figure}[h]
    \centering
    \includegraphics[width = 1.0\textwidth]{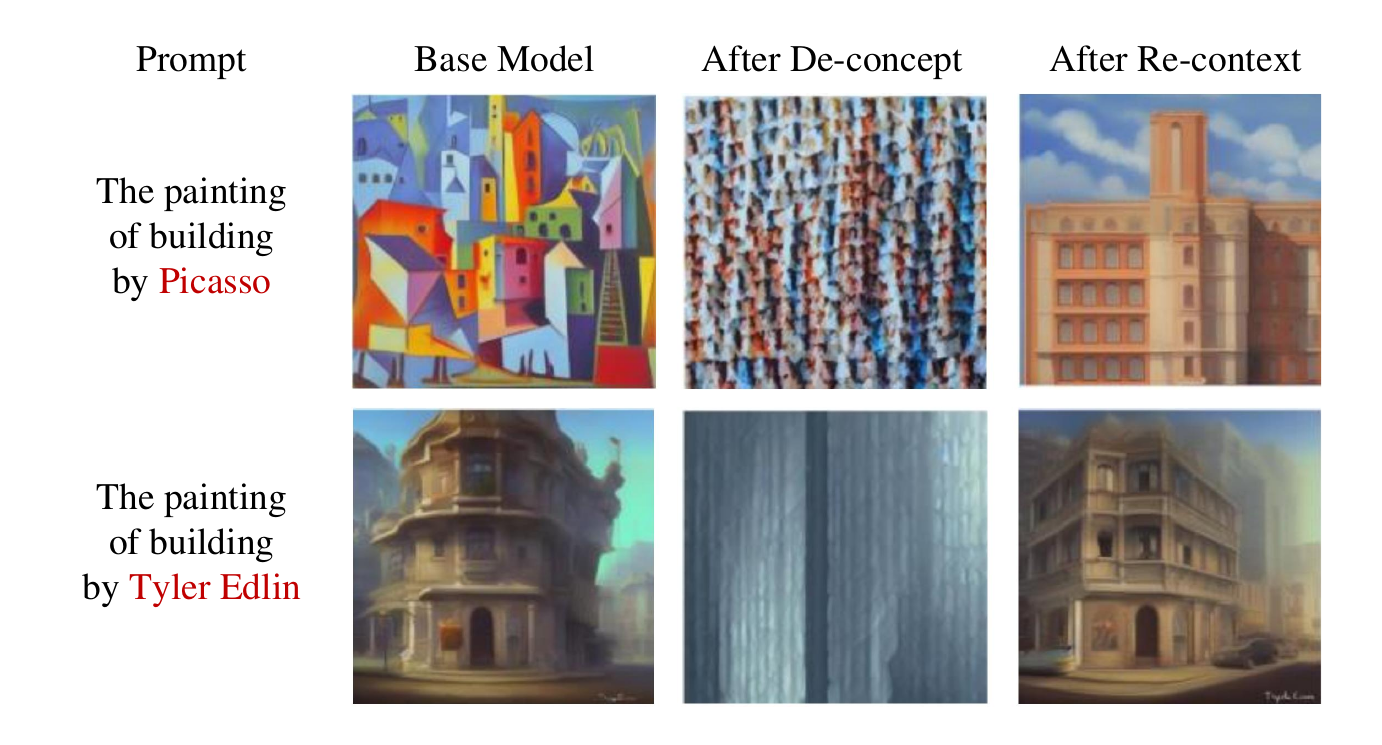}
    \caption{\textbf{Intermediate results of \emph{extraction}.} After the de-concept step, the non-infringing model's generative abilities become significantly limited, predominantly manifesting as the production of noise. 
    After the re-context step, the generation's prowess is rejuvenated, but due to the absence of learning Picasso-style images,  the model remains unable to generate artwork in the style of Picasso.}
    \label{fig:result of extraction}
\end{figure}

\subsection{Experiment on ordinary objects generation}\label{app:MScoco}
We evaluated the influence of \emph{extraction} on the generation of ordinary objects. We utilize 5000 textual captions selected from the validation set in MS-COCO \citep{lin2015microsoft} as prompts, generating 5000 images using SD1.5 and the non-infringing model that extracts R2D2 and Picasso, respectively. Several randomly selected images are displayed in Figure \ref{fig:results of unconditional sample extraction}. For illustrative purposes, we also generated results for concept-ablation and ESD on MS-COCO, respectively. Images within the same column exhibit substantial similarity, indicating that extraction does not exert an impact on the generation of ordinary items.

\begin{table}[!ht]
\caption{\textbf{Quantitative results on MS-COCO.} FID and KID metrics for removing the Picasso style are presented in the upper two rows, while those for removing R2D2  are displayed in the lower two rows. }
\label{tab:results of unconditional sample extraction}
\vskip 0.15in
\centering
\begin{tabular}{cccc}
\multicolumn{1}{c}{\bf Domain} &\multicolumn{1}{c}{\bf Method} &\multicolumn{1}{c}{\bf FID $\downarrow$} &\multicolumn{1}{c}{\bf {KID$\times 10^3$$\downarrow$}}  \\
\midrule
\multirow{2}{*}{Style replication} & \emph{Extract Picasso} & 24.04     & 2.83   \\
&\emph{Erase Picasso}    & 25.20     & 3.39 \\
\\
\multirow{2}{*}{IP recreation} &\emph{Extract R2D2} &   20.55      & 2.36  \\ 
&\emph{Ablate R2D2} & 18.97        & 1.34  \\

\end{tabular}
\vskip -0.1in
\end{table}

\begin{figure}[!ht]
    \centering
    \includegraphics[width = 0.9\textwidth]{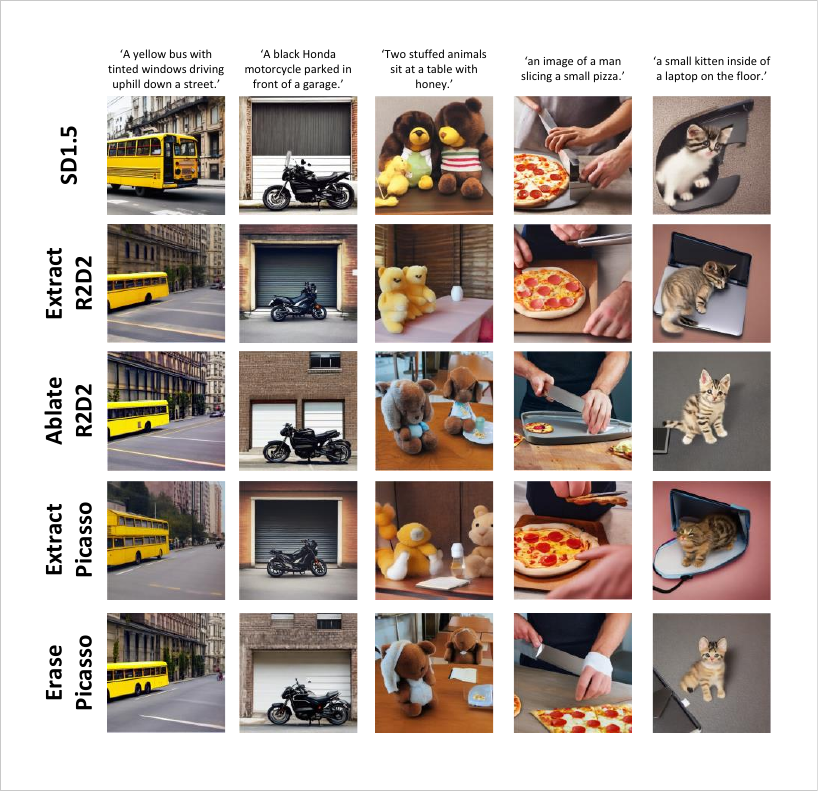}
    \caption{\textbf{Ordinary objects generation after \emph{extraction}.} Row 1 displays images generated by Stable Diffusion V-1.5. Rows 2 and 3 illustrate images generated after the removal of the IP character R2D2, while Rows 4 and 5 showcase images generated after the elimination of Picasso's style. Rows 3 and 5 serve as the baseline, representing concept-ablation and ESD, respectively. Notably, after the \emph{extraction} of R2D2 and Picasso, the non-infringing model retains the capability to generate commonplace objects sourced from the MS-COCO dataset \citep{lin2015microsoft}.}
    \label{fig:results of unconditional sample extraction}
\end{figure}

As depicted in Table\ref{tab:results of unconditional sample extraction}, we calculate some quantitative metrics like FID and KID. It is noteworthy that Concept-Ablation \citep{kumari2023ablating} only releases the checkpoint of ablating ``R2D2'' and ESD \citep{gandikota2023erasing} only releases the checkpoint of erasing ``Picasso''. Thus, we compare them separately by using their respective checkpoints. Evaluation metrics consistently maintain low values, further affirming that extraction does not compromise the generation of ordinary items.

\subsection{More Quantitative Results}

We compare the results of three methods following the removal of Van Gogh's influence. Like LPIPS, DINO-v2 \cite{oquab2023dinov2} is a self-supervised model used for extracting visual features from images, allowing us to compute the similarity between generated images generated by the non-infringing model and the base model. The higher the value of DINO-v2, the closer the image is.
Also, we calculate the CLIP-t distance to measure the semantic similarity between images generated by the non-infringing model and their corresponding textual prompts. The higher the CLIP-t value, the closer the sematic similarity is.

To ensure fairness in our evaluation, we implemented various settings to control the extraction effect and align it as closely as possible with the surrounding style. The quantitative results, as detailed in Table \ref{tab:Quantitative comparison of different methods}, demonstrate that our method achieves superior removal of the target style while preserving the integrity of the surrounding style. 

\definecolor{targetstyle}{RGB}{255, 0, 0} 
\definecolor{surroundingstyle}{RGB}{0, 0, 255} 
\begin{table}[h]
\centering
\caption{\textbf{Quantitative comparison of different methods.} The symbol $\uparrow$ indicates that the higher value is better on the metric, whereas $\downarrow$ symbol signifies the lower value is more preferable.The colors of symbols $\uparrow$ and $\downarrow$ should be read according to the colors of the most left column (Style).}
\label{tab:Quantitative comparison of different methods}

\begin{tabular}{@{}ccccccc@{}}

\textbf{Style} & \textbf{Methods} & \textbf{KID$\times 10^3$}\textcolor{targetstyle}{$\uparrow
$}\textcolor{surroundingstyle}{$\downarrow$} & \textbf{LPIPS}\textcolor{targetstyle}{$\uparrow
$}\textcolor{surroundingstyle}{$\downarrow$} & \textbf{DINO-v2$\%$}\textcolor{targetstyle}{$\downarrow
$}\textcolor{surroundingstyle}{$\uparrow$} & \textbf{CLIP-t$\%$}\textcolor{targetstyle}{$\downarrow
$}\textcolor{surroundingstyle}{$\uparrow$} \\ \midrule
\multirow{2}{*}{\textcolor{targetstyle}{\textbf{Target Style}}}   & ESD                        & 138          & 0.385          & 46.06            & 26.9           \\
               & EXTRACTION (OURS)                     & 187          & 0.387          & 38.8             & 23.9           \\[0.5\normalbaselineskip]
\multirow{2}{*}{\textcolor{surroundingstyle}{\textbf{Surrounding Style}}} & ESD                      & 27           & 0.212          & 71.42            & 32.15          \\
                  & EXTRACTION (OURS)                    & 32           & 0.157          & 79.02            & 31.9           \\[0.3\normalbaselineskip] \midrule 
\multirow{2}{*}{\textcolor{targetstyle}{\textbf{Target Style}}}   & CONCEPT-ABLATION           & 42           & 0.255          & 52.1             & 28.1           \\
               & EXTRACTION (OURS)                       & 58          & 0.293          & 48.6             & 28.2           \\[0.5\normalbaselineskip]
\multirow{2}{*}{\textcolor{surroundingstyle}{\textbf{Surrounding Style}}} & CONCEPT-ABLATION         & 12           & 0.123          & 81.2             & 29.6           \\
                  & EXTRACTION (OURS)                     & 7.4           & 0.115          & 86.2            & 32.8           \\[0.5\normalbaselineskip] 
\end{tabular}

\end{table}

\subsection{Experiment on seen and unseen contents generation}
In \emph{extraction}, we sample 10 common contents (training set) leveraging ChatGPT to fine-tune the base model. These contents have been previously processed by the non-infringing model. Additionally, we generate 10 supplementary contents for evaluation. Figure \ref{fig:seen&unseen} shows the images generated with these 20 contents. The images on the left represent the seen contents (training set), while those on the right are the unseen contents (evaluation set). Within each image block, we extract the corresponding artistic style.  The top row is generated by the base model, wheras the bottom row is generated by the non-infringing model. All images are generated with the prompt "[content] by [artist]". The specifics of the artistic and contents are detailed in Table \ref{tab:seen&unseen}.

\begin{figure}[!ht]
    \centering
    \includegraphics[width = \textwidth]{figures/results/seen-unseen.pdf}
    \caption{\textbf{Seen and Unseen contents generation after \emph{extraction}.} 
    On the left are 10 contents already seen during the \emph{extraction}, while on the right aren't seen. Within each image block, the top row is generated by the base model, and the non-infringing model generates the bottom one. Zoom in for better visualization. 
    }
    \label{fig:seen&unseen}
\end{figure}

\begin{table}[!ht]
\caption{\textbf{Details of artistic style and contents }}
\label{tab:seen&unseen}
\vskip 0.15in
\centering
\begin{tabular}{ccc}
Artistic Style & Seen Contents (training set) & Unseen Contents (evaluation set) \\
\midrule
Claude Monet & vase of flowers &  barn in a rural setting \\
Vincent van Gogh & bowl of fruit &  bustling street market \\
Pablo Picasso & still life with candles & boat on a calm river  \\
Johannes Vermeer & landscape with rolling hills &  group of animals in a field \\
Leonardo da Vinci & cityscape with buildings &  crowded café scene \\
Michelangelo & forest with sunlight filtering through trees & horse grazing in a pasture  \\
Frida Kahlo & portrait of a person & vintage clock on a mantelpiece  \\
Rembrandt van Rijn & quiet beach at sunset &  window with a view of the countryside \\
Salvador Dalí & mountain range with snow & room with antique furniture  \\
Henri Matisse & tranquil lake with reflections & close-up of a tree's bark and leaves  \\

\end{tabular}
\vskip -0.1in
\end{table}

To evaluate the performance of the base model and the non-infringing model, we computed various image distance metrics, including KID, LPIPS, and DINO-v2, between images generated by these two models. Additionally, we assessed the CLIP-t distance to measure the semantic alignment between textual prompts and images produced by the non-infringing model.

The quantitative results, averaged on the seen and unseen contents separately, are summarized in Table \ref{tab:Quantitative results on seen and unseen contents}. The lack of significant differences across all metrics for the seen versus unseen content suggests that the model did not exhibit signs of over-fitting.

\begin{table}[!ht]
\caption{\textbf{Quantitative results on seen and unseen contents.} The gap in metrics between seen content and unseen content is not significant. The symbol $\uparrow$ indicates that the higher value is better on the metric, whereas $\downarrow$ symbol signifies the lower value is more preferable. }
\label{tab:Quantitative results on seen and unseen contents}
\vskip 0.15in
\centering

\begin{tabular}{cccccc}
\multicolumn{1}{c}{\bf Style} &\multicolumn{1}{c}{\bf Contents} & \textbf{KID$\times 10^3$}\textcolor{targetstyle}{$\uparrow
$}\textcolor{surroundingstyle}{$\downarrow$} & \textbf{LPIPS}\textcolor{targetstyle}{$\uparrow
$}\textcolor{surroundingstyle}{$\downarrow$} & \textbf{DINO-v2$\%$}\textcolor{targetstyle}{$\downarrow
$}\textcolor{surroundingstyle}{$\uparrow$} & \textbf{CLIP-t$\%$}\textcolor{targetstyle}{$\downarrow
$}\textcolor{surroundingstyle}{$\uparrow$}\\
\midrule
\multirow{2}{*}{\textcolor{targetstyle}{\textbf{Target Style}}} & \emph{Seen Contents }   & 137   & 0.315 & 46.8 & 28.5 \\
&\emph{Unseen Contents }    & 168 & 0.303 & 43.3 & 28.0\\ [0.5\normalbaselineskip]
\multirow{2}{*}{\textcolor{surroundingstyle}{\textbf{Surrounding Style}}} &\emph{Seen Contents }      & 17.92 & 0.139 & 81.0 & 32.3 \\ 
&\emph{Unseen Contents }       & 16.66 & 0.135 & 81.6 & 32.1 \\

\end{tabular}
\vskip -0.1in
\end{table}

\subsection{Extracting specific prompt v.s. extracting style}

Figure \ref{fig:extract_prompt} compares the outputs  of the base model and the non-infringing model under two different extraction scenarios, extracting specific prmpt "sunflowers by van gogh" and entire style "van gogh". For each scenario presented, the left column exhibits images generated by the base model, whereas the right column features images produced by the non-infringing model.
The results indicate that the extraction method is effective not only in isolating and modifying individual content-style associations but also in comprehensively altering an artist's entire style. 

Table \ref{tab:Quantitative results on COCO caption set} shows that both extracting specific prompts and styles have little impact on the model's generative ability.

\begin{figure}[!ht]
    \centering
    \includegraphics[width = \textwidth]{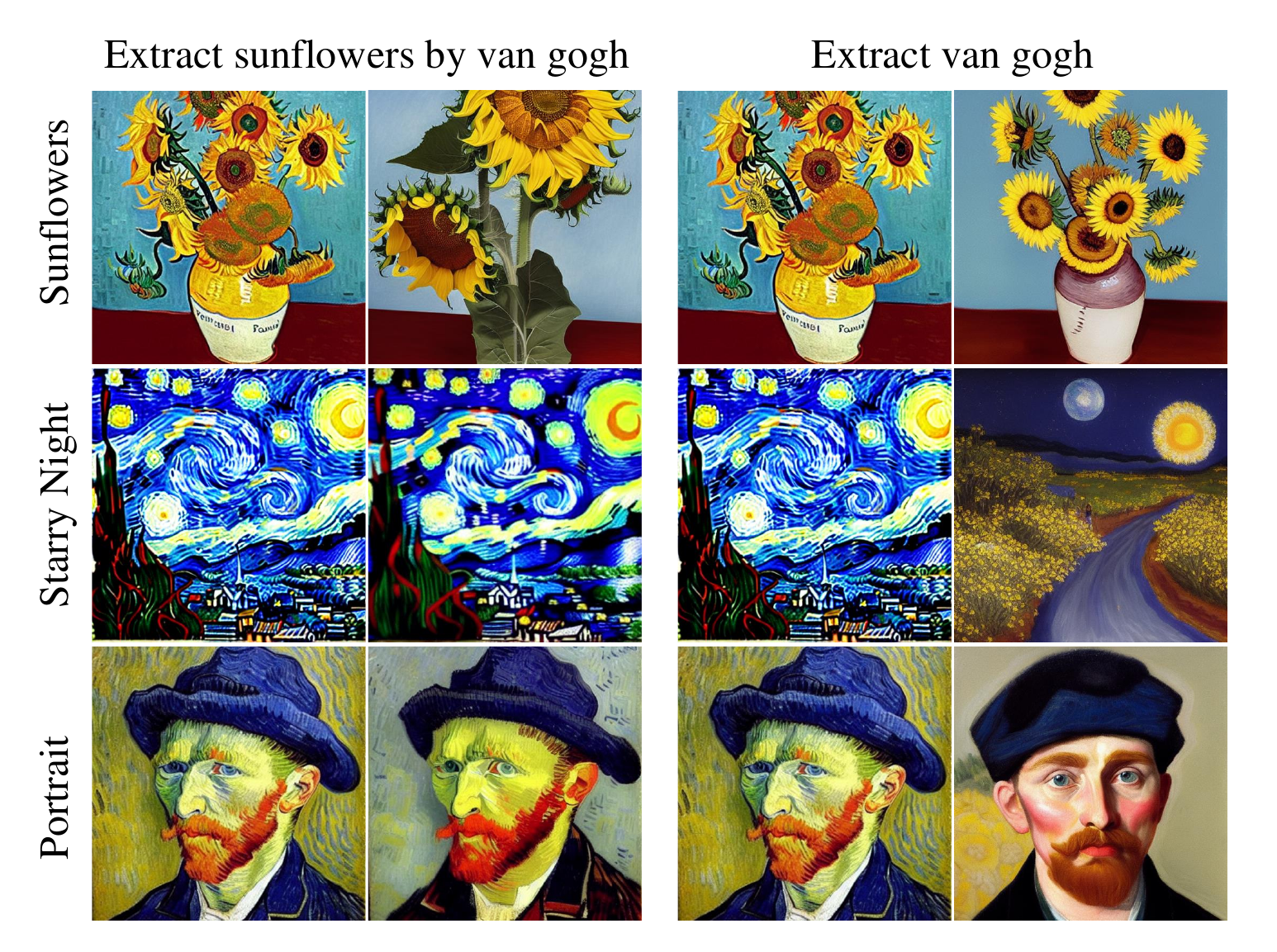}
    \caption{\textbf{The results of the non-infringing model after extracting specific prompt and style.} In the left block, only the specific prompt "sunflowers by van gogh" is extracted, while in the right block the entire style "van gogh" is extracted. For each block, the left column displays images generated by the base model, whereas the right column shows images generated by the corresponding non-infringing model.
    }
    \label{fig:extract_prompt}
\end{figure}

\begin{table}[h]
\centering
\caption{\textbf{Quantitative results on COCO caption set.}}
\label{tab:Quantitative results on COCO caption set}
\begin{tabular}{@{}llccl@{}}

\multicolumn{1}{l}{Method} & \multicolumn{1}{c}{} & \textbf{FID$\downarrow$} & \textbf{KID$\times 10^3$ $\downarrow$} \\ \midrule
Extract "van gogh"      &                      & 22          & 3.4          \\
Extract "sunflower by van gogh" &                      & 20.8        & 1.4          \\ 
\end{tabular}
\end{table}

\subsection{Style-IP \copyright plug-in combination}

Figure \ref{fig:IP-style_combination} shows the combination of style and IP \copyright plug-ins. The initial image is generated by the non-infringing model after extracting R2D2 and van Gogh, where the IP and style ate hard to recognize. Subsequent images, the second and third, are produced by the model after separately adding the R2D2 and van Gogh \copyright plug-in, thus distinctly featuring the respective IP and style. The final image is created after adding the combined \copyright plug-in, successfully achieving IP recreation and style replication.

\begin{figure}[!ht]
    \centering
    \includegraphics[width = \textwidth]{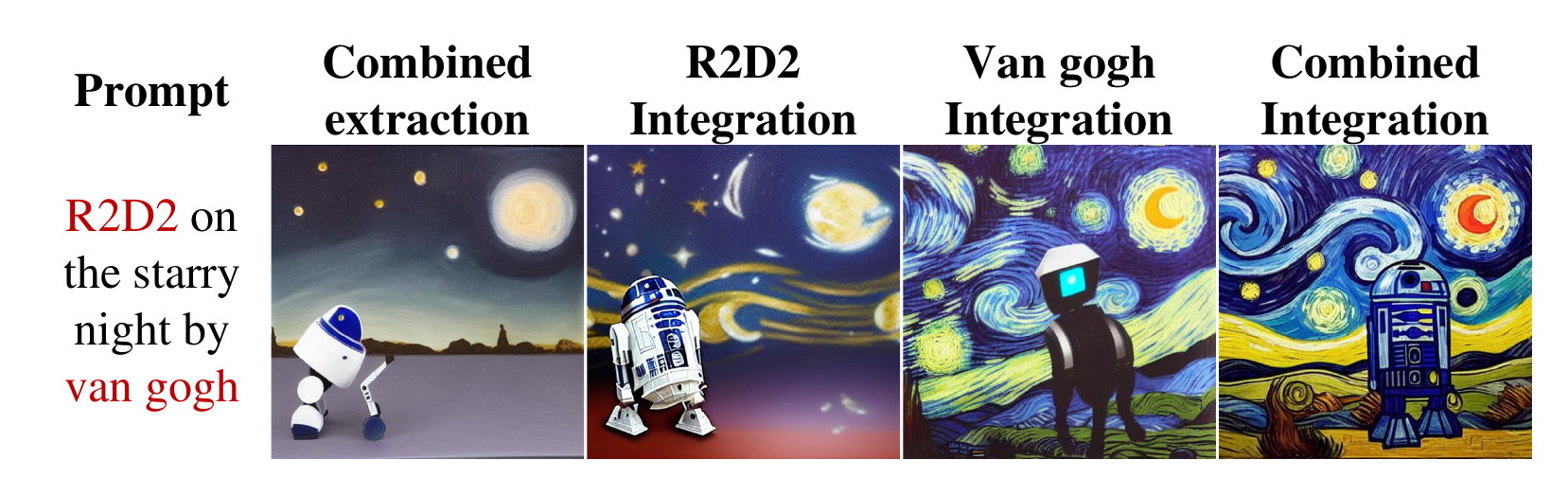}
    \caption{\textbf{IP Recreation and style replication in a single image.} We can integrate \copyright plug-in into the non-infringing model to recreate R2D2 or replicate van Gogh's style in a single image. Also, we can integrate the combined \copyright plug-in to achieve them both in an image.
    }
    \label{fig:IP-style_combination}
\end{figure}

\subsection{The limitation of potential degradation in the performance of the non-infringing model}

We observe a decline in the performance of the non-infringing model as the number of extracted styles increases. This degradation can be effectively mitigated by increasing the rate of re-context iterations and de-concept iterations, i.e., we can sample more contexts in the re-contexts in the re-context sub-process to maintain the generative ability of the non-infringing model.

During the \emph{extract} process, we set the number of de-concept iterations to be 10 and set the number of re-context iterations to $10\times r$. The default value for the rate is 1. 
For \emph{extraction}, we leverage ChatGPT to generate $10\times r$ common contents. 
For each iteration, we select one of these contents to generate 8 images. A higher $r$ allows the re-context phase to learn more contextual information, thereby maintaining the model's ability to generate high-quality images even as the number of extracted styles grows. 

Figure \ref{fig:limitation_of_extraction} illustrates this phenomenon. The first row displays the image generated by the base model. Subsequent rows, from top to bottom, are images generated by non-infringing models that extract 1 to 10 styles, respectively. As the number of extracted styles increases, there is a noticeable decline in image quality. However, by increasing the rate, this degradation is significantly alleviated, as evidenced by improved image fidelity.
The styles and contexts are detailed in Table \ref{tab:seen&unseen}. When the rate is greater than 1, we use the context in the evaluation set for the re-context phase.

\begin{figure}[!ht]
    \centering
    \includegraphics[width = \textwidth]{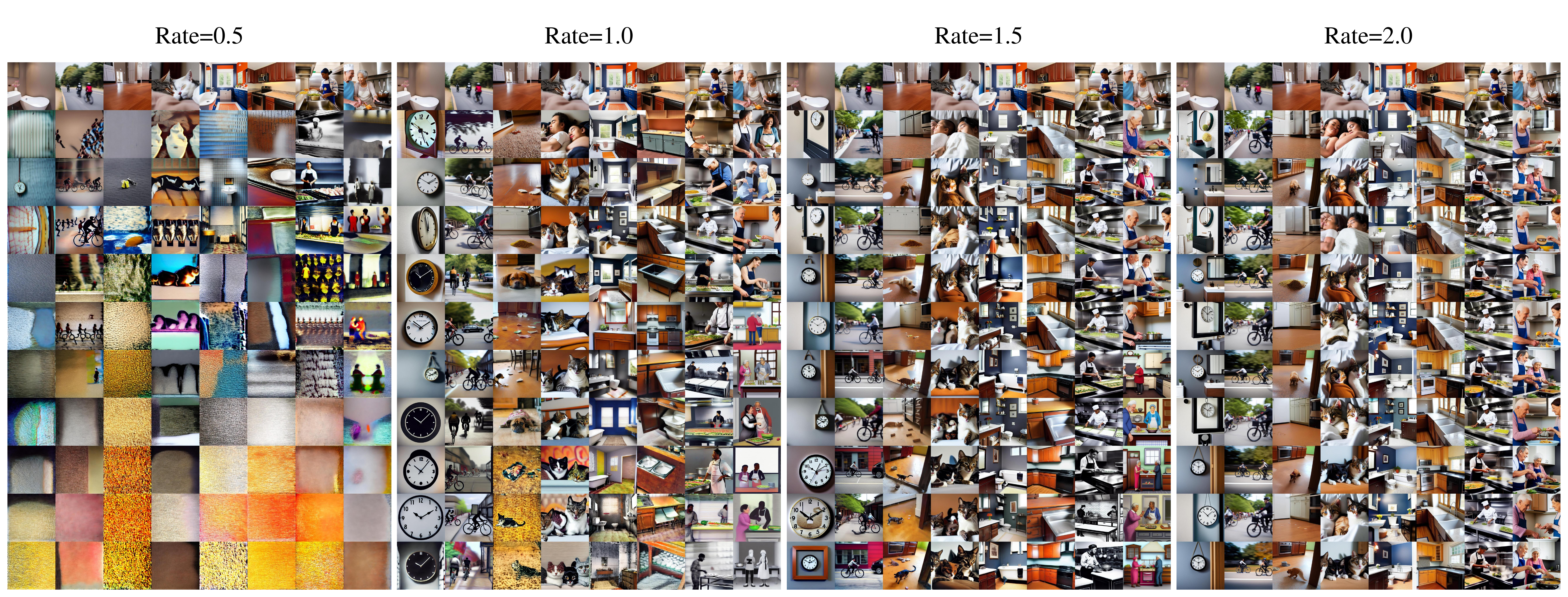}
    \caption{\textbf{The degradation and alleviation in the performance of the non-infringing model.} We randomly sample 8 images generated by different models. The first line is the image generated by the base model. The rest, from top to bottom, are images generated by non-infringing models that extract 1, 2, 3..., 10 styles, respectively. As the extracted concept number increases, the quality of images continuously declines. With the increase of the rate of re-context iteration number and de-concept iteration number, this decline has been effectively alleviated.
    Zoom in for better visualization.
    }
    \label{fig:limitation_of_extraction}
\end{figure}

\end{document}